\documentclass[conference]{IEEEtran}
\usepackage[T1]{fontenc}
\usepackage[utf8]{inputenc}
\DeclareUnicodeCharacter{2212}{\ensuremath{-}}
\DeclareUnicodeCharacter{00B1}{\ensuremath{\pm}}
\usepackage{amsmath,amssymb,booktabs,array,calc,graphicx}
\usepackage[table]{xcolor}
\definecolor{STVCgroup}{HTML}{ECEEEF}
\definecolor{STVCfocus}{HTML}{EDF7FA}
\usepackage{cite}
\let\citep\cite
\let\citet\cite
\usepackage{hyperref}
\hypersetup{hidelinks,pdftitle={PHASE: Pairing History and Anchoring Surgical Evidence for Vision-Language Video Understanding}}
\providecommand{\tightlist}{\setlength{\itemsep}{0pt}\setlength{\parskip}{0pt}}
\makeatletter
\@ifundefined{insert@pcolumn}{\let\insert@pcolumn\insert@column}{}
\newcommand{\STVCfloatstart}{\phantomsection\edef\STVCfloatanchor{\@currentHref}}
\newcommand{\STVCfloatreset}{\let\@currentHref\STVCfloatanchor}
\makeatother
\title{PHASE: Pairing History and Anchoring Surgical\\Evidence for Vision Language Video Understanding}
\author{
\IEEEauthorblockN{Ziyan Chen\textsuperscript{1*}, Sihang Zhao\textsuperscript{1*},
Taoyu Wu\textsuperscript{1\ensuremath{\dagger}}, Yiyi Miao\textsuperscript{1},\\
Zhengyong Jiang\textsuperscript{1}, 
Jionglong Su\textsuperscript{2 \ensuremath{\ddagger}},
Limin Yu\textsuperscript{1\ensuremath{\ddagger}},  and Zhe Tang\textsuperscript{1}}
\IEEEauthorblockA{\textsuperscript{1}Academy of Artificial Intelligence and Advanced Technology, Xi'an Jiaotong-Liverpool University, China\\
\textsuperscript{2}Institute of Artificial Intelligence Innovation, Zhejiang University of Technology, China}
}
\newcommand{\PHASEauthornotes}{%
  \begingroup
  \renewcommand{\thefootnote}{}%
  \footnotetext{%
    \textsuperscript{*}Contributed equally to this work.\par
    \textsuperscript{\ensuremath{\dagger}}Project leader.\par
    \textsuperscript{\ensuremath{\ddagger}}Corresponding authors.}%
  \endgroup
}
\begin{document}
\maketitle
\PHASEauthornotes
\begin{abstract}
Temporal experts can give surgical vision-language models (VLMs) phase evidence
distilled from preceding video, but this evidence is only as reliable as the history
behind it. Smoke, recording gaps and segment boundaries routinely leave that history
stale, mismatched or missing, and a corrupted history can mislead the expert more than
having no history at all. We observe that such failures never change the current clip
or its phase label. Building on this invariance, PHASE anchors training on each
clip--label pair and pairs true with disturbed histories at two stages. A
History-aware Phase Expert learns to predict the same label under both histories, and
the VLM is adapted on case-isolated expert scores that present every clip under both
conditions. On CholecT50, PHASE achieves the highest accuracy and case-averaged F1 among
temporal, expert-only and image-only VLM baselines, reaching 75.21\% F1 on clean and
71.33\% on synthetic-smoke inputs. Matched retraining identifies paired expert training
as the largest contributor and shows that per-window pairing outperforms globally
balanced exposure. The trained recognizer also keeps its predictions when history is
removed or replaced at inference. Anchoring on the invariant clip thus turns unreliable
history from a deployment hazard into a training signal.
\end{abstract}
\begin{IEEEkeywords}
Surgical phase recognition, vision-language models, temporal context modeling, robust learning.
\end{IEEEkeywords}

\section{Introduction}\label{introduction}

Surgical phase recognition assigns a procedural stage to the current moment of an
operation. A promising design pairs a temporal expert, which condenses preceding video
into phase scores \citep{endonet2016,tecno2020,mosformer2025}, with a surgical
vision-language model (VLM) that reads these scores alongside the current images
\citep{endochat2025,surgvlm2025,surgsage2026}. The expert's evidence, however, is only
as reliable as the history behind it. Smoke, recording gaps and segment boundaries leave
stale, mismatched or missing history. Replacing a clip's history with one from another
video drops an expert's macro-F1 from 43.39\% to 21.93\%, below the 36.96\% of a head
that ignores history (Table~\ref{tbl:interference}). Prior work improves how the
temporal model represents or filters history \citep{mosformer2025,dsted2026}, but how the
expert and the VLM should be \emph{trained}, so that the recognizer benefits from history
without depending on it, remains open.

Our key observation is that history failures never change the current clip or its phase
label. Each training window is therefore a fixed anchor around which true and disturbed
histories can be paired under identical supervision. This turns unreliable history from
a deployment hazard into a free training signal and raises the question this paper
answers: \emph{can clip-anchored history pairing, applied to both the temporal expert
and the VLM, yield a recognizer that exploits temporal evidence yet stays stable when
that evidence fails?}

To answer it, we propose \textbf{PHASE} (Pairing History and Anchoring Surgical
Evidence), which pairs histories at both training stages. History-paired interference
training (HIT) first supervises a History-aware Phase Expert (HPE) on true and disturbed
histories of the same clip. Complementary adaptation then shows the Qwen VLM
\citep{qwen354b} every window once with true-history and once with disturbed-history
scores from case-excluded experts \citep{stacking1992}, so every window, not just the
whole set, sees both conditions.

On the extended CholecT50 development group, PHASE reaches 75.21\% clean and 71.33\%
synthetic-smoke case-averaged F1 (CaseF1), the best among temporal, expert-only and
image-only VLM baselines. HIT is the largest contributor (+3.54/+3.60 points), and
without it the expert-conditioned VLM falls below the image-only VLM. When history is
removed or replaced at inference, PHASE keeps 208/210 and 210/210 predictions, whereas the
expert alone changes 26 and 50. Our contributions are:
\begin{itemize}
\tightlist
\item \textbf{Clip-anchored history pairing.} We show that the current clip and its
phase label are invariant to history failure and turn this invariance into a shared
training principle for a temporal expert and a surgical VLM.
\item \textbf{Pairing at both stages.} PHASE applies the principle twice: HIT keeps the
expert's prediction stable under disturbed history, and case-isolated complementary
adaptation shows every VLM training window with both true- and disturbed-history scores.
\item \textbf{Accurate and history-robust recognition.} PHASE achieves the best accuracy
and CaseF1 among the compared systems, matched retraining attributes the gains to pairing
at both stages, and the trained recognizer keeps its predictions when history fails at
inference.
\end{itemize}

\section{Related Work}\label{related-work}

\subsection{Temporal context and unreliable history}\label{temporal-context-and-unreliable-history}

Temporal phase recognition combines frame-level evidence with context from preceding
observations. TeCNO uses causal multi-stage temporal convolutions \citep{tecno2020},
Trans-SVNet fuses spatial and temporal embeddings \citep{transsvnet2021}, LoViT models
longer-range dependencies \citep{lovit2023}, TUNeS couples a temporal U-Net with
self-attention \citep{tunes2025}, and STAR-Net regularizes sequences of temporal
action-aware features \citep{starnet2023}. These architectures treat the history they
consume as informative. MoSFormer studies phase-history memory and cached impressions,
including inference-time history interventions \citep{mosformer2025}, and DSTED filters
unreliable historical features and refines ambiguous representations \citep{dsted2026}.
Both act on the architecture of the temporal model. PHASE acts on training: it holds the
current observation and label fixed and varies only the history, so that the expert
learns directly from history failure and the downstream VLM is adapted on the resulting
evidence.

\subsection{Expert evidence for surgical VLMs}\label{specialistvlm-interfaces}

Surgical video analysis spans endoscopic scene reconstruction and SLAM
\citep{endowave2025,shan20264d,shan2026nrgs,wu2025endoflow} as well as multimodal understanding. SurgVLP learns
transferable surgical representations from video lectures \citep{surgvlp2025}, and EndoChat and SurgVLM adapt multimodal models to surgical tasks
\citep{endochat2025,surgvlm2025}. VS-Assistant routes surgical requests to visual
analysis functions \citep{vsassistant2024}, and Surg-SAGE converts specialist predictions
into structured text for a frozen language model performing surgical visual question
answering \citep{surgsage2026}. These interfaces treat specialist outputs as given at
inference. When one model is trained on another model's predictions, however, the
provenance of those predictions matters: stacked generalization fits the second learner
on held-out predictions \citep{stacking1992}. PHASE extends this principle in two
directions. Training scores come from case-excluded experts, and each adaptation window
receives both true- and disturbed-history scores. LoRA \citep{lora2021} keeps the VLM
adaptation parameter-efficient.

\section{Method}\label{method}

\subsection{Overview and pairing principle}\label{system-architecture}

\begin{figure*}[!t]
\centering
\STVCfloatstart
\includegraphics[width=\textwidth]{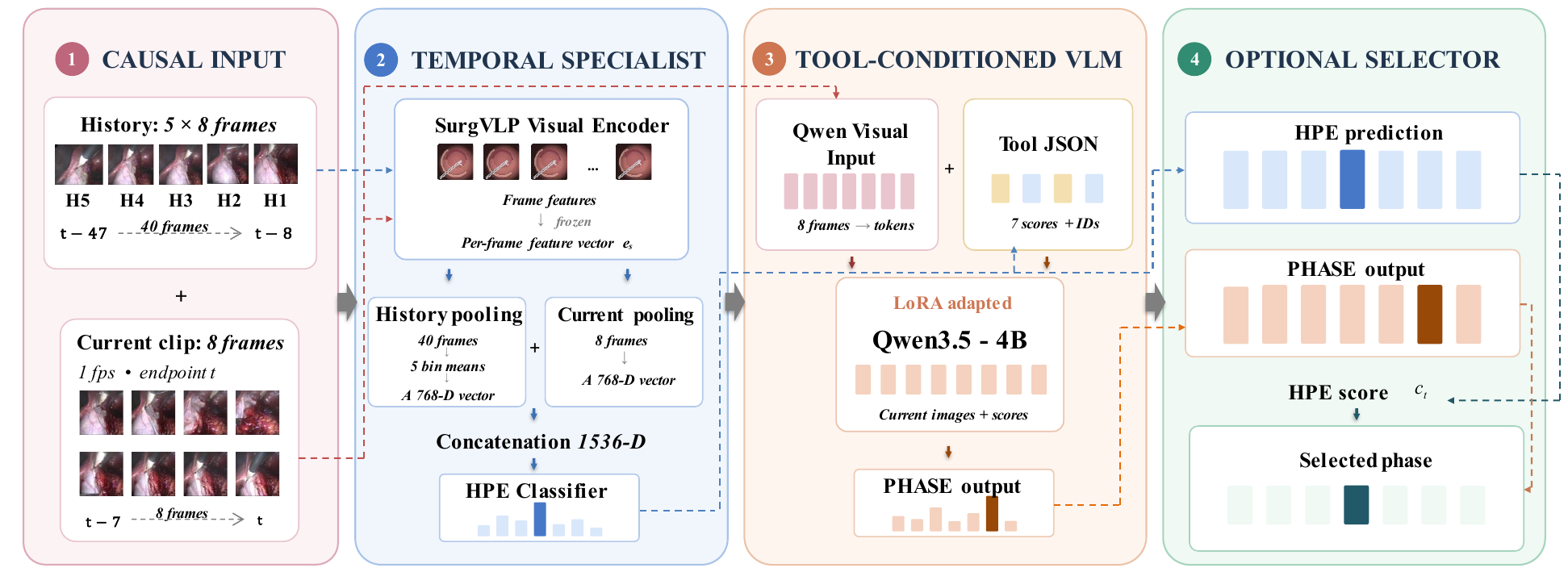}
\caption{PHASE overview. HPE pools current and historical SurgVLP features; Qwen
receives its phase scores alongside current images. Confidence-gated
selection uses HPE's maximum score \(c_t\) to choose between precomputed
predictions. The illustrated 40-frame history follows the valid-segment
rule for partial or empty bins. Highlighted output classes are
illustrative, not Qwen probabilities.}\STVCfloatreset\label{fig:architecture}
\end{figure*}

PHASE predicts the endpoint phase of an eight-frame clip from the current images and
temporal expert evidence (Fig.~\ref{fig:architecture}). At 1~fps, clip
\(X_t=(I_{t-7},\ldots,I_t)\) has target \(y_t\in\{0,\ldots,6\}\), and its history
consists of strictly preceding frames. The \textbf{History-aware Phase Expert (HPE)} maps
the clip and its history to phase scores, and an expert-conditioned Qwen VLM predicts
\(y_t\) from these scores and \(X_t\).

The history supplied to the expert may be true, stale, taken from another video or
missing. In every case the pair \((X_t,y_t)\) is unchanged. This is the basis of
\emph{clip-anchored history pairing}: PHASE uses the pair as the training anchor at both
stages. HIT supervises HPE on true and disturbed histories of the
same anchor (Section~\ref{history-paired-interference-training}), and case-isolated
complementary adaptation presents each anchor to the VLM once with true-history and once
with disturbed-history scores
(Section~\ref{case-isolated-complementary-vlm-adaptation}). We compare PHASE with
\textbf{HPE-only} and with a separately adapted, image-only \textbf{Qwen-only} branch.
PHASE results use the VLM output directly; the confidence-based selector in
Fig.~\ref{fig:architecture} serves as a rule-based fusion reference
(Section~\ref{optional-confidence-based-selection}).

\subsection{History-aware Phase Expert with paired supervision}\label{history-paired-interference-training}

\noindent\textbf{Separating current and historical evidence.} Pairing requires that a history
intervention leave the current representation untouched, so HPE pools the current clip
and its history separately. Frozen SurgVLP ResNet-50 encoder \(\phi\) and 768-dimensional
projection \(\psi\) \citep{surgvlp2025} give \(e_s=\psi(\phi(P_R(I_s)))\) for frame
\(s\), where \(P_R\) is SurgVLP's native preprocessing. If \(b_t\) starts the current
continuous valid segment, the five history bins are

\begin{equation}
\label{eq:history-bins}
\begin{aligned}&\mathcal B_{t,j}=\{s:\max(b_t,t-8j-7)\le s\le t-8j\},\\
&j=1,\ldots,5.\end{aligned}
\end{equation}

History is thus restricted to the current valid segment. A nonempty bin has validity
indicator \(m_{t,j}=1\) and mean feature \(h_{t,j}\); an empty bin has
\(m_{t,j}=0\) and \(h_{t,j}=0\). The current and historical summaries are

\begin{equation}
\label{eq:pooled-features}
x_t=\frac18\sum_{s=t-7}^{t}e_s,\qquad
\bar h_t=\frac{\sum_{j=1}^{5}m_{t,j}h_{t,j}}{\max(1,\sum_{j=1}^{5}m_{t,j})}.
\end{equation}

With \(x_t,\bar h_t\in\mathbb R^{768}\), the input \([x_t;\bar h_t]\in\mathbb R^{1536}\)
covers at most 48 frames and weights valid bins equally. Head \(H_\theta\) applies
LayerNorm \citep{layernorm2016}, a \(1536\rightarrow256\) linear map, GELU
\citep{gelu2016}, dropout 0.1 and a \(256\rightarrow7\) linear map:

\begin{equation}
\label{eq:hpe-prediction}
\begin{aligned}
p_t^{\mathrm{HPE}}&=\operatorname{softmax}(H_\theta([x_t;\bar h_t])),\\
\hat y_t^{\mathrm{HPE}}&=\arg\max_k p_{t,k}^{\mathrm{HPE}},\quad c_t=\max_k p_{t,k}^{\mathrm{HPE}}.
\end{aligned}
\end{equation}

Here \(p_t^{\mathrm{HPE}}\in\mathbb R^7\), \(k\) indexes phases, \(\hat
y_t^{\mathrm{HPE}}\) is HPE's prediction and the maximum softmax score \(c_t\) is an
uncalibrated selection statistic \citep{calibration2017}.

\noindent\textbf{History-paired interference training.} After a conventional fit of the
concatenated features, HIT continues the head with two supervised passes per example
that share the current summary and label. For a minibatch of \(M\) examples indexed by
\(i\), the head minimizes

\begin{equation}
\label{eq:hpe-loss}
\begin{aligned}
\mathcal L_{\mathrm{HPE}}(\theta)
&=\frac1M\sum_{i=1}^{M}\Bigl[
\tfrac12\operatorname{CE}(H_\theta([x_i;\bar h_i]),y_i)\\
&\qquad+\tfrac12\operatorname{CE}(H_\theta([x_i;\widetilde{\bar h}_i]),y_i)\Bigr],
\end{aligned}
\end{equation}

where \(\operatorname{CE}(z,y)=-\log[\operatorname{softmax}(z)]_y\) for logits \(z\).
The first pass uses true history. In the second, the summary
\(\widetilde{\bar h}_i\) replaces the history of a subset of examples with an older
(aged), a cross-training-video (wrong) or a zeroed (missing) history while preserving
\(x_i\) and \(y_i\); the remaining examples keep true history. Each intervention type
requests one sixth of the planned batch, rounded down. Donors exclude evaluation videos
and are drawn independently of recipient labels, so a cross-video history changes both
appearance and phase context. HIT updates the lightweight head alone and keeps the visual encoder frozen.

\subsection{Expert-conditioned visual input}\label{temporalvisual-dual-path-inference}

\begin{figure*}[!t]
\centering
\STVCfloatstart
\includegraphics[width=\textwidth]{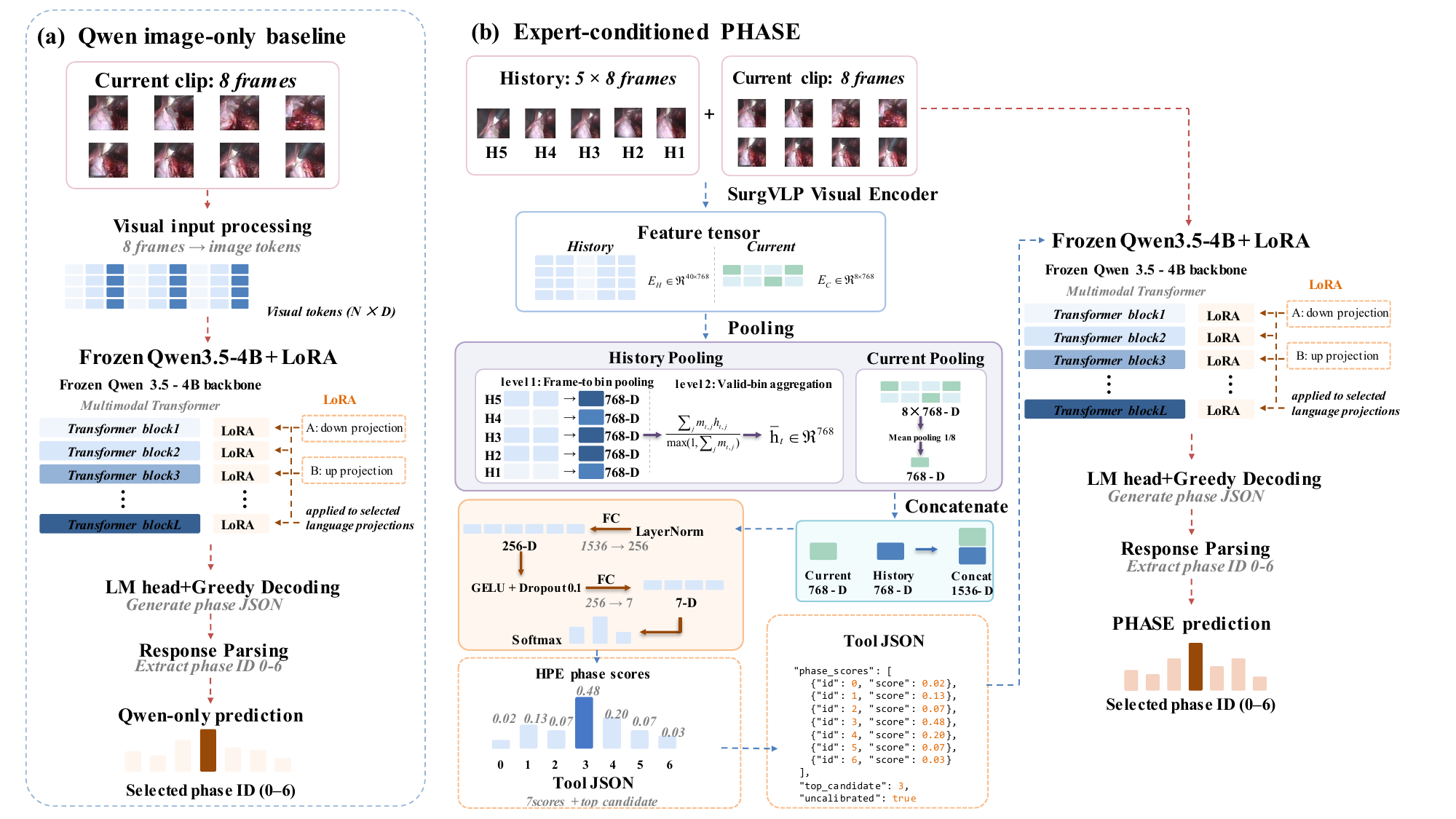}
\caption{Qwen-only and PHASE inference with frozen Qwen3.5-4B
\citep{qwen354b} and separate LoRA adapters \citep{lora2021}.
(a) Current images. (b) Images and HPE scores from frozen SurgVLP
\citep{surgvlp2025}, illustrated with 40 history frames. HPE scores are
illustrative; highlighted outputs denote classes, not VLM
probabilities.}\STVCfloatreset\label{fig:qwen-comparison}
\end{figure*}

The VLM receives HPE's evidence as a class-labelled summary next to the current images
(Fig.~\ref{fig:qwen-comparison}). Let \(P_{\mathrm{Qwen}}\) denote native Qwen3.5-4B
preprocessing \citep{qwen354b} and \(q\) the fixed seven-phase instruction. JSON
serialization \(J\) contains all seven HPE class scores, the top candidate and an
uncalibrated-score flag. The two branches receive

\begin{equation}
\label{eq:vlm-inputs}
\begin{aligned}
U_t^{\mathrm{Qwen}}&=(P_{\mathrm{Qwen}}(X_t),q,\mathrm{TOOL}=\mathrm{null}),\\
U_t^{\mathrm{PHASE}}&=(P_{\mathrm{Qwen}}(X_t),q,J(p_t^{\mathrm{HPE}})).
\end{aligned}
\end{equation}

Both inputs contain the current images, so a history change alters only the tool field.
The full score vector preserves alternative phases rather than committing the VLM to
the expert's top choice. The tool field carries predicted evidence alone, without ground truth, intervention
identity or SurgVLP features. Separate adapters decode the phase integers
\(\hat y_t^{\mathrm{Qwen}}\) and \(\hat y_t^{\mathrm{PHASE}}\).

\subsection{Case-isolated complementary adaptation}\label{case-isolated-complementary-vlm-adaptation}

\noindent\textbf{Case-isolated scores.} An expert scoring the cases it was fitted on produces
evidence that is more reliable than what it provides for unseen cases at deployment. Following
stacked generalization \citep{stacking1992}, we partition the training videos into three
folds, and window \(i\) in fold \(f_i\in\{1,2,3\}\) receives scores from
\(H_{\theta^{(-f_i)}}\). Fold \(f_i\) is excluded from both expert-training stages and
from the history donors. These out-of-fold (OOF) scores condition VLM adaptation, and
evaluation uses the HPE fitted on all training videos.

\noindent\textbf{Complementary exposure.} Each window receives true-history scores in one epoch
and corruption-matched wrong- or missing-history scores in the other, with its current
pixels and target fixed. Within each visual condition, every window draws an index
\(j\ge0\) from a shuffled order and keeps it across epochs. Epoch \(e\in\{0,1\}\) selects
entry \((j+2e)\bmod4\) of (real, real, matched-wrong, missing). The offset exchanges the
true and disturbed groups between epochs, so every window receives one condition from
each group. Qwen-only follows the same image and target schedule with a null tool field.

LoRA \citep{lora2021} augments a frozen weight \(W_0\in\mathbb R^{d_{\mathrm{out}}\times
d_{\mathrm{in}}}\) as

\begin{equation}
\label{eq:lora}
W'=W_0+\frac{\alpha}{r}B_L A_L,
\end{equation}

with trainable \(A_L\in\mathbb R^{r\times d_{\mathrm{in}}}\), \(B_L\in\mathbb
R^{d_{\mathrm{out}}\times r}\), \(r=8\), \(\alpha=16\) and zero dropout. Branch
parameters \(\eta^a\), \(a\in\{\mathrm{Qwen},\mathrm{PHASE}\}\), comprise full-attention
query/value and fused linear-attention query--key--value adapters; HPE and the Qwen base
weights (BF16) stay frozen, and adapters use FP32. Each branch minimizes, over \(N\)
exposures with answer length \(L_i\), tokens \(w_{i,\ell}\) and preceding tokens
\(w_{i,<\ell}\),

\begin{equation}
\label{eq:vlm-loss}
\begin{aligned}
\mathcal L_Q^a(\eta^a)
&=-\frac1N\sum_{i=1}^{N}\frac1{L_i}\sum_{\ell=1}^{L_i}\\
&\quad\log p_{\eta^a}(w_{i,\ell}\mid U_i^a,w_{i,<\ell}),
\end{aligned}
\end{equation}

where \(p_{\eta^a}\) is the next-token distribution. Prefix tokens are masked, and the
six answer tokens (phase JSON and end token) are weighted equally. The official chat
template runs with thinking disabled and native image grids.
At evaluation, greedy decoding (one beam, at most 16 tokens) generates the phase label
freely, and every reported prediction is the VLM's own answer: invalid or truncated JSON
counts as an error, with no fallback to HPE.

\subsection{Confidence-gated selection}\label{optional-confidence-based-selection}

The selector in Fig.~\ref{fig:architecture} implements a rule-based alternative to
learned conditioning in the spirit of selective prediction
\citep{selectivenet2019,defer2020}. It defers to HPE whenever HPE is confident:

\begin{equation}
\label{eq:selection}
\hat y_t^{\mathrm{sel}}(\tau)=\begin{cases}\hat y_t^{\mathrm{HPE}},&c_t\ge\tau,\\\hat y_t^{\mathrm{PHASE}},&c_t<\tau,\end{cases}
\end{equation}

with threshold \(\tau\). It operates offline on precomputed HPE and PHASE predictions.
Section~\ref{sec:deployment} compares it with ungated PHASE.

\section{Experimental}\label{experimental}\label{experimental-results}

The experiments follow the three contributions. Sections~\ref{quantitative-comparison}
and~\ref{qualitative-comparison} establish accurate recognition,
Section~\ref{ablation-study} attributes it to pairing at both stages, and
Section~\ref{sec:deployment} tests robustness when history fails, from the expert to
the full recognizer.

\subsection{Implementation Details}\label{implementation-details}

\noindent\textbf{Experimental Setup.} The three OOF heads fit 23/23/24 videos with 690
conventional and 345 HIT updates (AdamW \citep{adamw2019}, learning rate 0.001, weight decay
0.0001, optimizer reset between stages). The deployment HPE fits all 35 training videos
and uses the development-selected HIT checkpoint; HPE-only reports this
head. VLM adaptation runs two epochs over 6,802 windows in two visual conditions (27,208
exposures, 13,604 updates) with microbatch one, gradient accumulation two, AdamW at
learning rate \(10^{-4}\), zero weight decay and gradient clipping at one. Qwen-only,
PHASE and all ablation variants share the random state, initial adapter weights, image
order, target encoding, prompts and decoding, and are evaluated at the epoch-two
checkpoint. As temporal baselines, current-only and compact TeCNO \citep{tecno2020} are
matched-endpoint controls trained on the same windows and visual conditions as the VLMs.
Current-only averages eight frozen SurgVLP features \citep{surgvlp2025}; compact TeCNO
processes six ordered pooled tokens over at most 48 frames with a causal two-stage head
(eight layers, 32 channels per stage). Both train for 405 updates over three runs
(AdamW, learning rate 0.001, weight decay 0.0001, batch 512). We further retrain the
phase-only ResNet50--TCN TeCNO \citep{tecno2020} and adapt Trans-SVNet
\citep{transsvnet2021} and TUNeS \citep{tunes2025} causally on frozen SurgVLP features
with their original temporal architectures. Trans-SVNet trains its TeCNO teacher for 8
epochs and its fusion module for 25; TUNeS trains for 75 epochs with sparse multi-scale
supervision, unmasked reconstruction and temporal smoothing, since its dense-label
components do not apply to endpoint labels. Both select checkpoints by
clean accuracy on the initial group. All baselines are evaluated at the shared endpoints.

\noindent\textbf{Datasets.}\label{datasets} We use local seven-phase annotations of
CholecT50 videos, distinct from its public action-triplet labels \citep{rendezvous2022}.
Phases 0--6 are preparation, Calot-triangle dissection, clipping and cutting,
gallbladder dissection, gallbladder packaging, cleaning and coagulation, and gallbladder
extraction. Each 1-fps sample contains eight valid current frames and an endpoint target;
no future images or labels are used. The 35/5/10 video partition follows Fold~1 of the official CholecT50 cross-validation split \citep{cholect50splits2022}. Training comprises 6,802 windows from 35 videos at
\(t=7+10k\). Two development groups, disjoint from the training videos, contain 5
(initial) and 10 (extended) videos with 765 and 2,113 windows per condition. The larger
extended group serves as the primary benchmark, and the initial group, which also guided
development, serves as a replication. Clean and synthetic-smoke views
(severity 2/3) \citep{corruptions2019} share targets and are evaluated separately,
giving 5,756 inputs across 15 videos.

\noindent\textbf{Evaluation Metrics.}\label{evaluation-metrics} Accuracy pools windows.
CaseF1 averages fixed-seven-class macro-F1 over \(n_V\) videos, weighting every case and
phase equally, with phase-wise counts \(TP_{v,k},FP_{v,k},FN_{v,k}\):

\begin{equation}
\label{eq:case-f1}
\operatorname{CaseF1}=\frac{1}{7n_V}\sum_{v=1}^{n_V}\sum_{k=0}^{6}
\frac{2TP_{v,k}}{2TP_{v,k}+FP_{v,k}+FN_{v,k}}.
\end{equation}

Undefined class F1 is set to zero, and an invalid answer counts as a false negative for
the true phase. Case-level uncertainty uses 2,000 paired whole-video bootstrap draws,
which preserve within-video dependence. Differences are computed before
rounding. The expert-level studies in Tables~\ref{tbl:context} and~\ref{tbl:interference}
report macro-F1 with global class counts.

\begin{table*}[!t]
\centering
\caption{Phase recognition on the extended CholecT50 development group \citep{rendezvous2022} (\%). Asterisks mark study-defined controls or adaptations; three-run rows report mean accuracy and CaseF1 mean±SD. Bold marks the best result per column.}\label{tbl:collaboration}
\small
\setlength{\tabcolsep}{3pt}
\renewcommand{\arraystretch}{1.12}
\begin{tabular}{@{}>{\raggedright\arraybackslash}p{(\linewidth-10\tabcolsep)*\real{0.27000}}>{\raggedleft\arraybackslash}p{(\linewidth-10\tabcolsep)*\real{0.06000}}>{\raggedleft\arraybackslash}p{(\linewidth-10\tabcolsep)*\real{0.14000}}>{\raggedleft\arraybackslash}p{(\linewidth-10\tabcolsep)*\real{0.19500}}>{\raggedleft\arraybackslash}p{(\linewidth-10\tabcolsep)*\real{0.14000}}>{\raggedleft\arraybackslash}p{(\linewidth-10\tabcolsep)*\real{0.19500}}@{}}
\toprule
\textbf{Method} & \textbf{Runs} & \multicolumn{2}{c}{\textbf{Clean}} & \multicolumn{2}{c}{\textbf{Smoke}} \\
\cmidrule(lr){3-4}\cmidrule(l){5-6}
 & & Acc$\uparrow$ & CaseF1$\uparrow$ & Acc$\uparrow$ & CaseF1$\uparrow$ \\
\midrule
Current-only\(^{*}\) \citep{surgvlp2025} & 3 & 65.55 & 55.50±0.86 & 63.56 & 54.93±1.20 \\
Compact TeCNO\(^{*}\) \citep{tecno2020} & 3 & 67.72 & 59.33±1.39 & 65.31 & 58.76±0.75 \\
TeCNO (phase-only) \citep{tecno2020} & 1 & 77.76 & 68.82 & 73.83 & 68.12 \\
Trans-SVNet\(^{*}\) \citep{transsvnet2021} & 1 & 75.11 & 63.93 & 72.36 & 63.91 \\
TUNeS\(^{*}\) \citep{tunes2025} & 1 & 70.28 & 59.13 & 69.52 & 59.57 \\
HPE-only\(^{*}\) \citep{surgvlp2025} & 1 & 71.42 & 60.23 & 59.96 & 45.66 \\
Qwen-only\(^{*}\) \citep{qwen354b} & 1 & 83.48 & 73.64 & 81.07 & 69.28 \\
\rowcolor{STVCfocus}
\textbf{PHASE (Ours)} & 1 & \textbf{83.77} & \textbf{75.21} & \textbf{81.26} & \textbf{71.33} \\
\bottomrule
\end{tabular}
\end{table*}

\begin{figure}[!t]
\centering
\STVCfloatstart
\includegraphics[width=\linewidth]{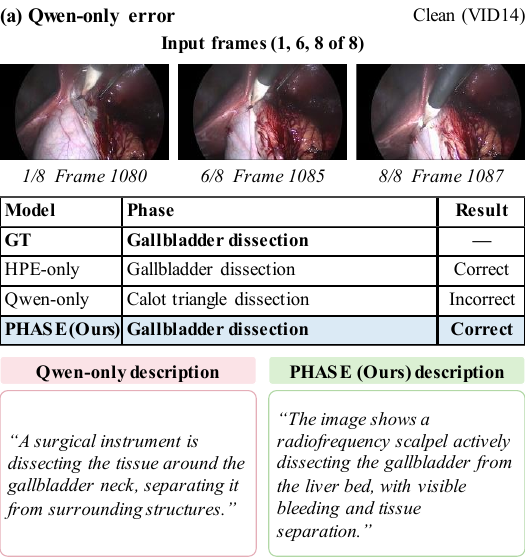}\par\smallskip
\includegraphics[width=\linewidth]{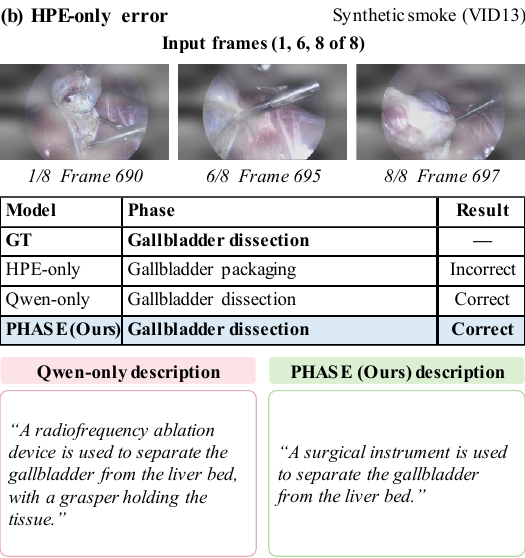}
\caption{Complementary corrections on selected development inputs: PHASE matches GT in both panels. (a) Qwen-only errs; HPE-only is correct. (b) HPE-only errs; Qwen-only is correct. Frames 1, 6 and 8 are shown; GT denotes ground truth. Additional-prompt descriptions are reproduced verbatim with JSON wrappers removed; labels match original predictions. They are separate from the original phase-prediction outputs. Colors identify models.}\STVCfloatreset\label{fig:qualitative}
\end{figure}

\subsection{Quantitative Comparison}\label{quantitative-comparison}

Table~\ref{tbl:collaboration} shows that PHASE achieves the highest accuracy and CaseF1 in
both visual conditions, with 75.21\% clean and 71.33\% smoke CaseF1. It exceeds the
strongest temporal baseline, phase-only TeCNO, by 6.39/3.21 CaseF1 points
(clean/smoke) and the expert it reads, HPE-only, by 14.98/25.66 points. The margin over
HPE-only widens under smoke, where the VLM's direct view of the current images
compensates for degraded temporal evidence: PHASE corrects 561 HPE-only errors while
introducing 111, a net gain of 450 correct smoke predictions.

Relative to the image-only Qwen-only branch, PHASE raises CaseF1 by 1.57/2.04 points and
accuracy by 0.28/0.19 points. The larger CaseF1 gain shows that the added temporal
evidence benefits phases and cases that pooled accuracy underweights. Case-level
bootstrap intervals for this CaseF1 difference are [−1.69, 4.98] (clean) and
[−0.54, 4.59] (smoke), and on the initial group the two branches obtain 73.05\%/70.61\%
(PHASE) and 74.36\%/69.59\% (Qwen-only) clean/smoke CaseF1. Because the two branches
differ in both input interface and training evidence, we attribute the gains through
matched retraining (Section~\ref{ablation-study}), which changes one factor at a time.

\subsection{Qualitative Comparison}\label{qualitative-comparison}

Fig.~\ref{fig:qualitative} shows the two complementary error patterns that PHASE
resolves. On clean VID14, Qwen-only mistakes gallbladder dissection for Calot-triangle
dissection, while the history-aware expert is correct, and PHASE agrees with the
expert. On smoke-degraded VID13, HPE-only predicts gallbladder packaging, while the
image-only VLM is correct, and PHASE agrees with the visual reading. In both cases PHASE
recovers the ground-truth phase, combining the strengths of the two evidence sources
rather than inheriting either one's errors.

\subsection{Ablation Study}\label{ablation-study}\label{expert-preparation}\label{adaptation-effects}\label{training-and-inference}\label{sec:expert-analysis}

Table~\ref{tbl:components} retrains the VLM with one component of PHASE changed at a
time. \textbf{HIT} denotes history-paired interference training, \textbf{Scores} the
expert-score input, \textbf{CI} case isolation and \textbf{Exposure} the history
schedule during adaptation; dashes denote inapplicability. All four variants share
PHASE's adaptation schedule, initial states and update budget.

\begin{table*}[!t]
\centering
\caption{Matched-retraining ablation on the extended CholecT50 development group \citep{rendezvous2022} (\%). Exposure: complementary (Comp.), real-only (Real) or random mixing (Rand.). Bold marks the best result per column.}\label{tbl:components}
\small
\setlength{\tabcolsep}{4pt}
\renewcommand{\arraystretch}{1.12}
\begin{tabular*}{\textwidth}{@{\extracolsep{\fill}}lccccrrrr@{}}
\toprule
\textbf{Variant} & \textbf{HIT} & \textbf{Scores} & \textbf{CI} & \textbf{Exposure} & \multicolumn{2}{c}{\textbf{Clean}} & \multicolumn{2}{c}{\textbf{Smoke}} \\
\cmidrule(lr){6-7}\cmidrule(l){8-9}
 & & & & & Acc$\uparrow$ & CaseF1$\uparrow$ & Acc$\uparrow$ & CaseF1$\uparrow$ \\
\midrule
HPE-only & \(\checkmark\) & --- & --- & --- & 71.42 & 60.23 & 59.96 & 45.66 \\
Qwen-only & --- & \(\times\) & --- & --- & 83.48 & 73.64 & 81.07 & 69.28 \\
\midrule
w/o HIT & \(\times\) & \(\checkmark\) & \(\checkmark\) & Comp. & 79.79 & 71.67 & 77.61 & 67.72 \\
w/o CI & \(\checkmark\) & \(\checkmark\) & \(\times\) & Comp. & 82.44 & 72.85 & 80.08 & 71.18 \\
Real-only exposure & \(\checkmark\) & \(\checkmark\) & \(\checkmark\) & Real & 83.06 & 72.96 & 81.07 & 70.98 \\
Random mixing & \(\checkmark\) & \(\checkmark\) & \(\checkmark\) & Rand. & 82.39 & 73.44 & 81.21 & \textbf{71.80} \\
\rowcolor{STVCfocus}
\textbf{PHASE (Ours)} & \(\checkmark\) & \(\checkmark\) & \(\checkmark\) & Comp. & \textbf{83.77} & \textbf{75.21} & \textbf{81.26} & 71.33 \\
\bottomrule
\end{tabular*}
\end{table*}

\noindent\textbf{Paired expert training is the largest factor.} Removing HIT lowers CaseF1 from
75.21\%/71.33\% to 71.67\%/67.72\%, a drop of 3.54/3.60 points and the largest effect in
Table~\ref{tbl:components}. The no-HIT control keeps both supervised passes, labels, optimization
budget, OOF split and deployment checkpoint, and replaces only the disturbed histories
with true ones. Without HIT, the expert-conditioned VLM falls below Qwen-only in all four
columns. Expert evidence therefore improves the VLM only when the expert has been
trained on history failure.

\noindent\textbf{Case isolation.} Case-isolated scores raise CaseF1 in both conditions,
most strongly on clean inputs (+2.36 points), and raise accuracy in both conditions
(83.77\% vs.\ 82.44\% and 81.26\% vs.\ 80.08\%). The no-CI control lets each receiving case be scored by an expert that has
seen it, through deterministic case swaps that preserve the 23/23/24 fitting cases and
the per-step label, augmentation and interference counts. Both settings use the same
deployment HPE at evaluation, so the comparison isolates how adaptation scores are
constructed.

\noindent\textbf{Per-window pairing outperforms global balance.} Real-only exposure supplies
true-history scores in both epochs. Random mixing preserves the global real/disturbed
totals but gives 1,686 windows true history twice, 1,686 disturbed history twice and
only 3,430 one of each. Complementary exposure gives all 6,802 windows one true- and one
disturbed-history exposure. With HPE heads, images, targets and budget fixed,
complementary exposure improves clean CaseF1 by 2.26 points over real-only and 1.77 over
random mixing, and attains the highest accuracy in both conditions. Under smoke, the
three schedules lie within 0.82 CaseF1 points (70.98\%--71.80\%) and all exceed Qwen-only
(69.28\%); the per-window guarantee pays off most on clean inputs.

\noindent\textbf{Joint effect and replication.} With HIT, every score-conditioned variant
exceeds Qwen-only in smoke CaseF1 (70.98\%--71.80\% vs.\ 69.28\%), and on clean inputs the
full combination of HIT, case isolation and complementary exposure lifts PHASE above it
(75.21\% vs.\ 73.64\%). The ranking replicates on the initial group, where full PHASE
(73.05\%/70.61\% clean/smoke CaseF1) exceeds all four retrained variants in both
conditions.

\noindent\textbf{Expert representation.} Table~\ref{tbl:context} compares
temporal representations for the expert head. At 345 updates, separate pooling reaches
61.15\%/40.93\% clean/degraded macro-F1, improving over current-only by 5.31/4.52
points, over joint pooling of all 48 frames by 4.39/3.55 and over compact TeCNO by
2.06/1.35. Compact TeCNO needs twice the updates to match its clean macro-F1 (61.19\% at
690 updates). Separate pooling also makes the pairing well defined, because history can
be replaced while \(x_t\) stays identical. This three-run study uses ten development
videos and clean-validation-selected checkpoints; degraded macro-F1 averages five
corruption families (smoke, motion blur, Gaussian blur, occlusion and brightness) at
three severities.

\begin{table}[!t]
\centering
\caption{Temporal representations for the expert head on CholecT50 \citep{rendezvous2022} (macro-F1, \%). Bold marks the best result at the shared 345-update budget.}\label{tbl:context}
\small
\setlength{\tabcolsep}{3pt}
\renewcommand{\arraystretch}{1.12}
\begin{tabular}{@{}>{\raggedright\arraybackslash}p{(\linewidth-6\tabcolsep)*\real{0.44000}}>{\raggedleft\arraybackslash}p{(\linewidth-6\tabcolsep)*\real{0.14000}}>{\raggedleft\arraybackslash}p{(\linewidth-6\tabcolsep)*\real{0.21000}}>{\raggedleft\arraybackslash}p{(\linewidth-6\tabcolsep)*\real{0.21000}}@{}}
\toprule
\textbf{Method} & \textbf{Updates} & \multicolumn{2}{c}{\textbf{Macro-F1}} \\
\cmidrule(l){3-4}
 & & Clean$\uparrow$ & Degraded$\uparrow$ \\
\midrule
Current-only\(^{*}\) \citep{surgvlp2025} & 345 & 55.84 & 36.41 \\
Joint pooling\(^{*}\) \citep{surgvlp2025} & 345 & 56.76 & 37.38 \\
\rowcolor{STVCfocus}
Separate pooling\(^{*}\) \citep{surgvlp2025} & 345 & \textbf{61.15} & \textbf{40.93} \\
Compact TeCNO\(^{*}\) \citep{tecno2020} & 345 & 59.09 & 39.58 \\
\midrule
Compact TeCNO\(^{*}\) \citep{tecno2020} & 690 & 61.19 & 43.32 \\
\bottomrule
\end{tabular}
\end{table}

\begin{table}[!t]
\centering
\caption{Expert macro-F1 under history interventions on CholecT50 \citep{rendezvous2022} (\%, mean and SD over two runs). Degraded columns average five corruption families.}\label{tbl:interference}
\scriptsize
\setlength{\tabcolsep}{2pt}
\renewcommand{\arraystretch}{1.15}
\newcommand{\sd}[1]{\,{\tiny±#1}}
\begin{tabular}{@{}lrrrrr@{}}
\toprule
\textbf{Method} & \textbf{Clean} & \multicolumn{4}{c}{\textbf{Degraded}} \\
\cmidrule(l){3-6}
 & Real & Real & Aged & Wrong & Missing \\
\midrule
Current-only\(^{*}\) & 55.15\sd{0.61} & 36.96\sd{0.56} & 36.96\sd{0.56} & 36.96\sd{0.56} & 36.96\sd{0.56} \\
Separate pooling\(^{*}\) & 64.37\sd{1.99} & 43.39\sd{0.86} & 42.31\sd{0.97} & 21.93\sd{0.45} & 34.62\sd{0.10} \\
Normal continuation\(^{*}\) & 62.29\sd{2.62} & 42.96\sd{1.26} & 41.72\sd{1.44} & 21.99\sd{0.96} & 32.69\sd{0.63} \\
History dropout\(^{*}\) & 62.55\sd{2.01} & 43.19\sd{0.26} & 41.83\sd{0.36} & 23.15\sd{0.42} & 39.03\sd{1.16} \\
\rowcolor{STVCfocus}
\textbf{HPE (HIT)}\(^{*}\) & 61.87\sd{3.49} & 42.09\sd{1.13} & 42.08\sd{1.18} & 30.23\sd{2.64} & 38.00\sd{1.08} \\
\bottomrule
\end{tabular}
\end{table}

\subsection{Robustness to History Failure}\label{sec:deployment}

We test robustness to failing history at two levels, first for the expert and then for
the full recognizer at inference.

\noindent\textbf{Expert-level robustness.} Table~\ref{tbl:interference} stresses
the expert with altered histories on five development videos over two runs. Normal
continuation, history dropout and HIT each extend the separate-pooling head by 345
updates; clean inputs use real history, degraded columns average the five corruption
families, and interventions unavailable for a window fall back to real history. Wrong history is the most damaging failure: it lowers separate pooling from
43.39\% to 21.93\%, below the history-free head (36.96\%). Normal continuation leaves
this failure intact (21.99\%), and history dropout, which simulates only missing history,
barely changes it (23.15\%). HIT raises wrong-history macro-F1 to 30.23\%, recovering
8.30 points, the largest gain among the continuations. Under missing history, HIT and
dropout both improve over normal continuation (38.00\% and 39.03\% vs.\ 32.69\%), while
real- and aged-history F1 under degradation stay within 1.3 points of separate pooling.
HIT thus shifts the expert's capacity from the clean, true-history case (61.87\% vs.\
64.37\%) toward the failures it will meet in deployment, and Table~\ref{tbl:components}
shows that this shift carries through to end-to-end recognition.

\noindent\textbf{Recognizer-level stability.} We hold the trained PHASE adapter fixed and
corrupt history at inference. Seven endpoints from each of the 15 development videos
are selected by phase coverage and then temporal coverage before any prediction is
inspected, giving 210 paired clean/smoke inputs. Missing history uses
zero \(5\times768\) features with an all-one mask; wrong history uses a training-video
donor matched by clean-feature cosine similarity, bin lengths and corruption condition.
Current features, images, prompts, preprocessing and greedy decoding stay fixed. Missing and wrong histories change 26
and 50 HPE predictions but only 2 and 0 PHASE predictions: the expert's instability does
not propagate to the final recognition. The recognizer also
degrades gracefully when the expert is unavailable. Removing the score field or replacing
it with uniform scores changes at most three of 210 answers, keeps 161--164 correct
answers against 163 for normal input, and yields valid outputs throughout.

\noindent\textbf{Learned conditioning versus confidence gating.} The selector in
Eq.~\eqref{eq:selection} replaces learned fusion with a fixed rule. On the extended
group, \(\tau=0.50\) lowers accuracy from 83.77\% to 74.16\% on clean inputs and from
81.26\% to 63.37\% under smoke, because confident HPE errors override correct VLM
predictions. Raising \(\tau\) restores smoke accuracy monotonically (79.32\% at 0.90,
80.50\% at 0.95, 81.16\% at 0.99), and the best thresholds recover ungated PHASE to within
0.1 points in both conditions; on the initial group, no threshold exceeds ungated PHASE. Rule-based deferral therefore adds nothing beyond what the
VLM learns from paired scores, and PHASE is deployed ungated.

\vspace{10pt}
\section{Conclusion and Discussion}\label{conclusion}

History fails during surgery; the current clip and its phase label do not. PHASE turns
this invariance into \emph{clip-anchored history pairing}, a training principle for
expert-conditioned phase recognition. Applying the \emph{pairing at both stages}, it
supervises the temporal expert on true and disturbed histories of the same clip and adapts
the VLM on case-isolated scores that present every clip under both conditions. The result
is \emph{accurate and history-robust recognition}: the highest CaseF1 and accuracy among
the compared systems on CholecT50, paired expert training as the largest single factor
(+3.54/+3.60 CaseF1), and predictions that hold when history is removed or replaced at
inference.

\vspace{3pt}
The ablations also give practical guidance, because each component controls a different
property of the evidence the VLM learns from: separate pooling fixes the current clip, HIT
shapes the expert's response to failing history, case isolation sets which cases the
expert has seen, and complementary exposure sets which history conditions accompany each
window. Together they give the clean-input advantage over the image-only VLM.

\vspace{3pt}
Our results also clarify where expert evidence acts in an expert-conditioned VLM.
Changing how the scores are prepared during training moves CaseF1 by up to 3.60 points,
whereas perturbing them at inference leaves the trained recognizer nearly unchanged.
Paired training thus lets the adapter absorb temporal evidence without hinging on any
single score vector, and learned conditioning makes rule-based deferral to the expert
unnecessary. Because it reuses existing phase labels and leaves the inference interface
unchanged, clip-anchored pairing is a drop-in recipe for any expert--VLM pipeline whose
context can fail while the observation and target stay fixed, including tool, anatomy and
workflow experts for surgical VLMs.

\bibliographystyle{IEEEtran}
\bibliography{references}
\end{document}